\documentclass[conference]{IEEEtran}
\IEEEoverridecommandlockouts
\usepackage[american]{babel}
\usepackage[utf8]{inputenc}
\usepackage{flushend}
\usepackage{graphicx}
\usepackage{amsfonts,amssymb,amsmath}
\usepackage{amsthm}
\usepackage{hyperref}
\usepackage{subcaption} 
\usepackage{subfloat}
\usepackage{color}
\usepackage{grffile}
\usepackage{tikz}
\usepackage{float}
\usepackage{verbatim}
\usepackage{algorithm}
\usepackage{algpseudocode}

\usepackage{marginnote}
\usepackage{acronym}

\makeatletter
    \newcommand{\linebreakand}{%
      \end{@IEEEauthorhalign}
      \hfill\mbox{}\par
      \mbox{}\hfill\begin{@IEEEauthorhalign}
    }
    \makeatother

\begin{document}

\title{Analyzing Echo-state Networks Using Fractal Dimension\\
\thanks{This work was financially supported by the Advanced Institute of Manufacturing with High-tech Innovations (AIM-HI) from The Featured Areas Research Center Program within the framework of the Higher Education Sprout Project by the Ministry of Education (MOE) in Taiwan.}
}

\author{\IEEEauthorblockN{Norbert Michael Mayer}
    \IEEEauthorblockA{\textit{Department of Electrical Engineering and} \\ \textit{Advanced Institute of Manufacturing with High-tech Innovations} \\
    \textit{National Chung Cheng University}\\
    Taiwan \\
    nmmayer@gmail.com}
    \and
    \IEEEauthorblockN{Oliver Obst}
    \IEEEauthorblockA{\textit{Centre for Research in Mathematics and Data Science} \\
    \textit{Western Sydney University}\\
    New South Wales \\ Australia \\
    O.Obst@westernsydney.edu.au}
}

\acrodef{ESN}{Echo State Network}
\acrodef{RNN}{Recurrent Neural Network}
\acrodef{RC}{Reservoir Computing}
\acrodef{MGS}{Mackey Glass System}
\acrodef{FD}{fractal dimension}
\acrodef{ESP}{echo state property}
\acrodef{SVM}{support vector machine}

\maketitle
%\IEEEpeerreviewmaketitle
\begin{abstract}
%% What does the work do?
This work joins aspects of reservoir optimization, information-theoretic optimal encoding, and at its center fractal analysis. 
%% Why is that useful/a good idea/interesting to do?
We build on the observation that, due to the recursive nature of recurrent neural networks, % the set of 
input sequences appear as fractal patterns in their hidden state representation. These patterns have a fractal dimension that is lower than the number of units in the reservoir. We show potential usage of this fractal dimension with regard to optimization of  % reservoir features or 
recurrent neural network initialization. 
%% How does it do that?
We connect the idea of ``ideal'' reservoirs to lossless optimal encoding using arithmetic encoders. 
%% What are the results?
Our investigation suggests that the fractal dimension of the mapping from input to hidden state shall be close to the number of units in the network. 
%% What is the significance of this? (e.g., does this make the world a better place, can we now do something useful we couldn't do before, ...)
%While we restrict the set of input values to be finite, the  
This connection between fractal dimension and network connectivity is an interesting new direction for recurrent neural network initialization and reservoir computing.
\end{abstract}


\section{Introduction}

\IEEEPARstart{O}{NE} of the 
key problems for \ac{RNN} initialization, and in particular for \ac{RC} methods like \acp{ESN}\cite{jaeger2001echo,jaeger2002short,jaeger2002tutorial} is that it is still unclear what kind of connectivity results in the best performance. 
For recurrent neural networks, some heuristics have shown much better performance than others: Dependent on the task, \acp{ESN} with orthogonal recurrent connectivity matrices that scale with the size of the network have shown better performance than other strategies~ \cite{WLS04,mayer2017orthogonal,boedecker2012information}. An initialization method that can create a good recurrent weight matrix for a given task would be a compromise between an arbitrarily chosen connectivity matrix, and a connectivity matrix found by more expensive optimization, e.g., by backpropagation through time. Insights that lead to better connectivity matrices are useful for ``standard'' \acp{RNN}, where they are further trained, as well as for \ac{RC} approaches, where they are used without further training.

With this in mind, the purpose of our work is to investigate how the input statistics is mapped to the \acp{ESN} reservoir. It appears self-evident that the performance of the network relates to an (ideally) optimal usage of the memory capacity of the recurrent layer neurons. Of the two related questions about the memory capacity (1) ``How much memory does a specific task require'', and (2) ``How much memory does a network of a given size provide'', we are addressing the latter. It should be clear that some tasks require more memory capacity than others, and that, beyond a certain point, simply increasing memory will not contribute to a better performance. We are interested in making the best possible use of a given network, but the question of how large that network should be, for a given task, is outside the scope of this paper.

In order to approach this optimal capacity of networks with a given size, we follow the concepts of lossless memory compression. 
Our results reveal that the recursive nature of the update function in \acp{RNN}, and in \acp{ESN} in particular, give rise to the occurrence of fractals with regard to certain features of the reservoir in a natural way. Suggesting the fractal dimension as a measure for reservoir quality, we discuss new directions for how to improve initialization and performance of \acp{ESN}.

To provide better insights into properties of optimal coding recurrent neural networks, we restrict our investigation to use a discrete set of two input values that drive a network, which incidentally also allows a very natural application of (Shannon) information theory.

\section{Mathematical background and definitions}

In this section, we briefly revisit the \acp{ESN} formalism, and some of the background for fractal dimensions and lossless compression.

\subsection{Echo State Networks}

An \ac{ESN} is described by the following equation:
\begin{align}
x_{t+1}  = \sigma (\alpha W x_t + \beta w_\text{in} u_t),
\label{eq:esn}
\end{align}
where $\sigma$ is a (non-linear) transfer function for which Lipschitz continuity is required, $u_t$ is an external input to the network, i.e., the stimulus. 
The $m$ recurrent neurons are represented by $x_t$, an $m$ dimensional vector. We set appropriate norms on both matrices $\|W\|_2=\|w_\text{in}\|_2=1$. The scalar values of $\alpha$ and $\beta$ can be used for tuning the network and their quantitative analysis.

To prevent divergence, \acp{ESN} should satisfy the so-called \ac{ESP} \cite{yildiz2012re,buehner2006tighter}. %  
Heuristics show that, for some tasks, performance of \acp{ESN} are best near the limit of what is permissible according to the \ac{ESP}.  Likewise, 
several theoretical works point in this direction \cite{mayer2004echo,boedecker2012information,hajnal2006critical}. 

Input values $u_t$ form an input stream (for example from sensory input, where the ESN forms or is embedded into a larger system). These values usually follow certain restrictions and form a statistics, which can be described in a likelihood relation, 
\[p(u_t) \equiv p(u_t|u_{t-1}, u_{t-2} \dots ).\]
One possible way to describe such statistics is to consider the set $\mathbb{U}$, where each possible time series $u_{(-\infty,t]}$ is in $\mathbb{U}$ which 
may be combined with the relative probability of its occurrence $p(u_{(-\infty,t]})$, in order to make the description of the statistics mathematically accurate.

Information propagation in \acp{RNN} can be viewed as a recursive process. We can describe an internal state of the network as $x_t = f(x_0,u_{[0,t]})$. Due to the \ac{ESP}, it is also that the state  
$x_t = f(u_{(-\infty,t]})$, i.e., the current internal state is independent of the initial state, if the network has received sufficiently long input sequences. 
Viewed differently, the state vector $x_t$ is set into a state space $\mathbb{X}$. Dimensionality of this space is equal to the number of hidden layer neurons. With a bounded transition function like $\sigma=\tanh$, output values of recurrent neurons are restricted to $[-1,1]$, and thus 
$\mathbb{X}$ forms a hypercube.

We restrict the input $u$ to the \ac{RNN}  so that $u_t \in \{-1,1\}$. In this case, the numerical results can be used to visualize interesting characteristics. In the following, we focus on the fact that the representations of the entire set $\mathbb{U}$ -- within one experimental setting (i.e., with certain statistics) onto the internal state-set $\mathbb{X}$ -- reveals fractals as overall phenomenon. In general, \acp{ESN} will usually have 100 or more units. For visualization, we confined the number of neurons to two neurons. The representation of $\mathbb{U}$ onto the internal state neurons reveals the fractal nature of the resulting point clouds for the naked eye (cf.\ Fig.~\ref{fig:1}).

\subsection{Fractal dimension and lossless compression}

Mathematician Benoit Mandelbrot in 1975 firstly used the term \emph{fractal} (``broken'', or ``fractured'') to apply the theoretical concept of \acp{FD} to geometric patterns in nature. Generally speaking, fractals are wrinkled objects that, to some extent, escape more conventional measures such as length and area, and are better distinguished by their \ac{FD}. The geometry of fractals comes up in many mathematical models for different objects in nature -- such as mountains, coastal lines and clouds~\cite{pentland1984fractal,peitgen2006chaos}. One way to think about \ac{FD} is based on the concept of self-similarity, i.e., the relationship between the number of identically shaped smaller objects the original object can be divided into, and their masses. For perfect self-similarity, this relationships can be described as
 \begin{align*}
 d_f =\lim_{r\to0}{\frac{\log (N_r)}{\log {\frac{1}{r}}}}
 \end{align*}
where $N_r$ is the least number of distinct copies of the original object $A$ at scale $r$. The union of $N_r$ distinct copies must cover $A$ completely. 
In general, fractals are not necessarily (perfectly) self-similar, but the \ac{FD} is a useful attribute of fractals in nature and some fields of natural sciences, in addition to their properties described by Euclidean geometry. 

Different methods have been proposed to calculate and estimate the \ac{FD}. Work in \cite{balghonaim1998maximum} classifies the methods of evaluating \acp{FD} into three main groups: (1) box-counting methods, (2) variance methods, and (3) spectral methods. In particular box-counting methods became very popular for their simplicity and auto-computability~\cite{peitgen2006chaos}, and are applied in various fields. A number of different box-counting methods has been brought forward~\cite{gagnepain1986fractal,sarkar1994efficient,chen2003two,novianto2003near}. 

 \newcommand{\de}{4.2cm}

 \begin{figure*}[t]
     \centering
     \begin{tabular}[t]{ccc}
     $\alpha=0.83$ &
     $\alpha=0.95$ with a magnified inset &
     $\alpha=0.99$ \\
      \includegraphics[width=\de]{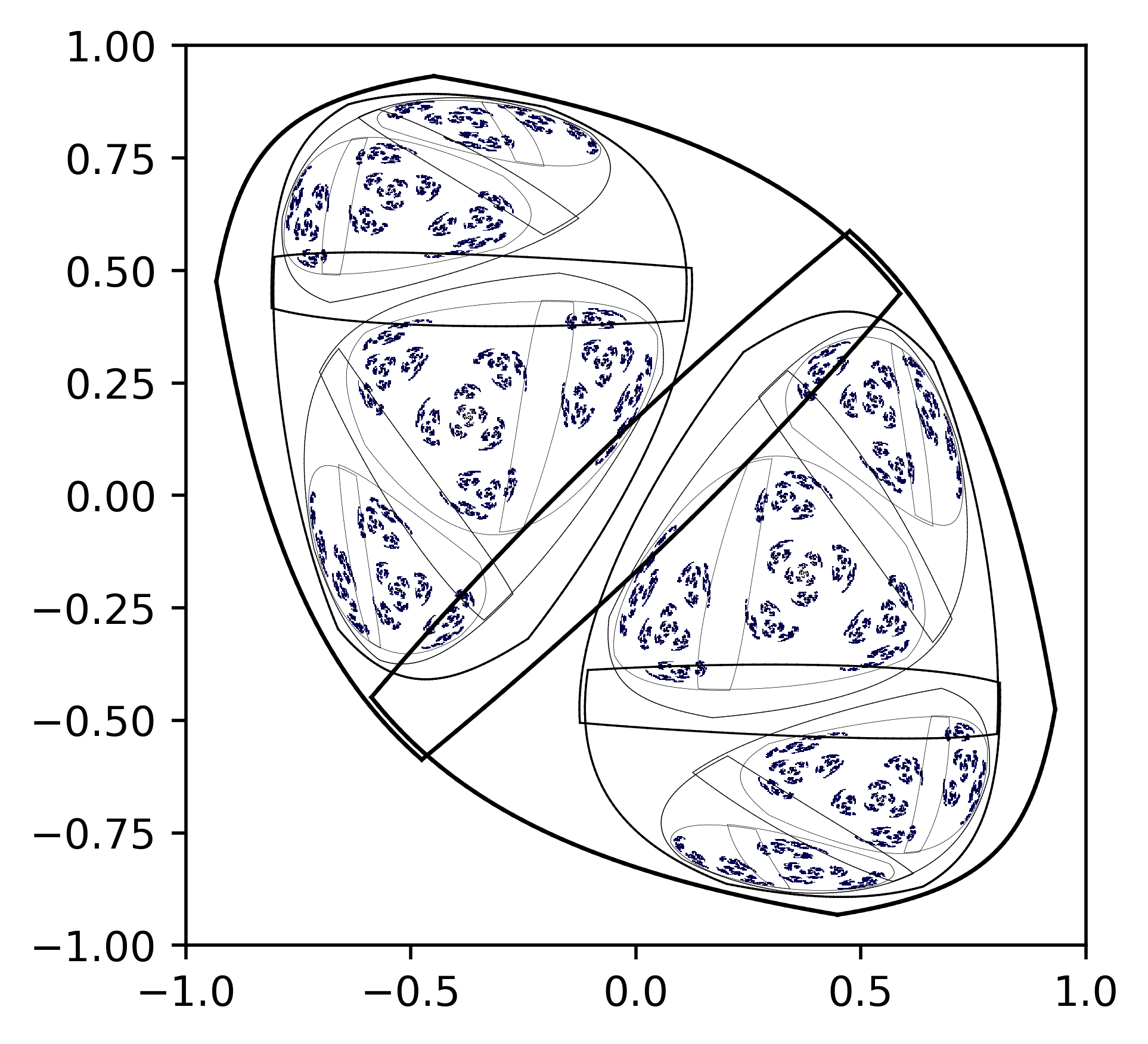} &
      \includegraphics[width=8.4cm]{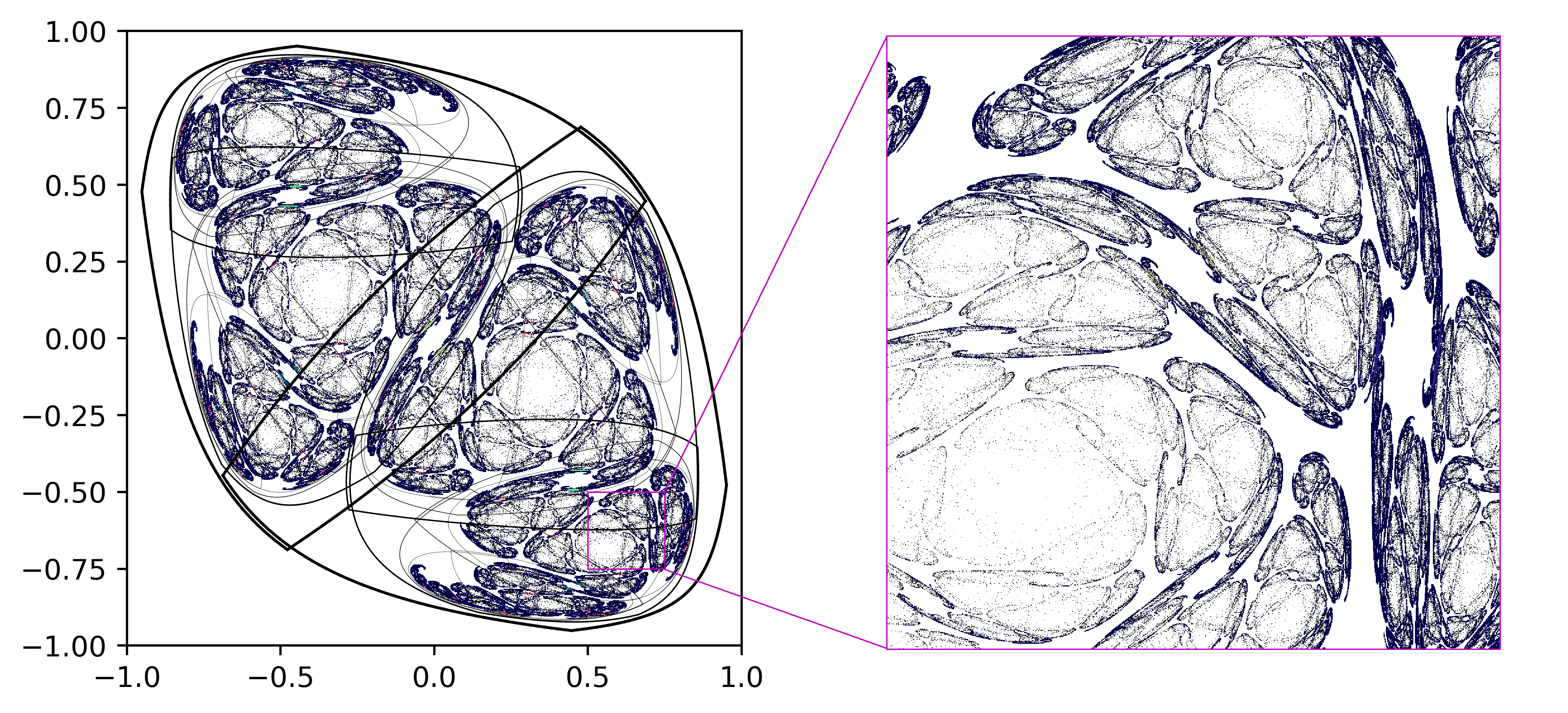} &
      \includegraphics[width=\de]{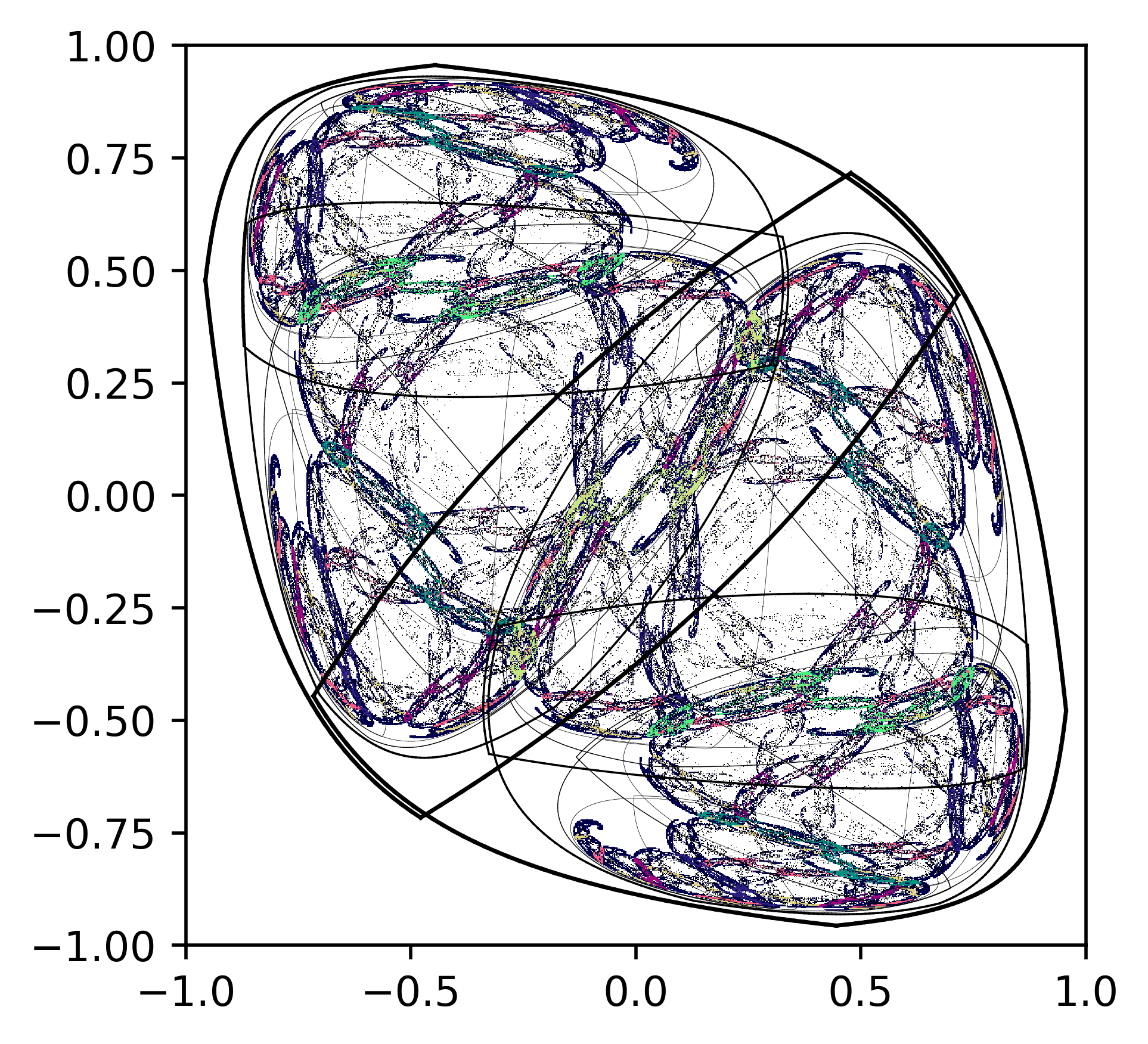} 
     \end{tabular}

     \caption{State space representation of an \ac{ESN}'s recurrent layer with 2 neurons. The closed black curves depict the transformed state space limits.}
 \label{fig:0}
 \end{figure*}

\section{RNNs and their fractal dimension}

As we investigate the mapping of the complete input statistics onto the reservoir manifold, one can see that representations of input histories emerge as fractal sets for many parameter settings of the connectivity matrix. Essentially, these fractals are an indicator of the `coarseness' of the occupied states in the underlying reservoir. Here, one can understand coarseness as a scale-invariant feature. In combination with the total number of states that the reservoir can assume, one can calculate the memory capacity of the reservoir for a given input statistics. The resulting fractal dimension are key components to the better information-theoretic understanding of reservoir computing and \ac{RNN} initialization. We discuss connections between optimal encoding in \acp{ESN} and arithmetic encoders, Shannon information in reservoirs, and implications for other types of networks.

Though popular due to their efficient training, \acp{ESN} also have some basic limitations \cite{jaeger2007echo}. A first limitation are potentially ill-conditioned solutions, resulting in models that might diverge from the real system that is supposed to be represented. This, and other limitations have also been observed in other works, e.g.,  \cite{jaeger2005reservoir,xue2007decoupled,steil2007online,song2010stable}, and improvements have been suggested in~\cite{liu2009improved,obst2010iconip,sheng2012prediction}. A better use of the available memory would alleviate some of these limitations.

One basic insight at the start of our research was that an optimal representation of states in a reservoir has an analogue in the representation pattern of arithmetic encoders \cite{rissanen1979arithmetic}.
Arithmetic encoders have capabilities comparable to Huffman codes: they encode a sequence of symbols in to a real valued number, that can then be transmitted with a given accuracy.
We use them as a guideline that can lead us to an optimal representation of the input for a given statistics on the hyperspace of the recurrent layer neurons' activities. It has been outlined in \cite{BC2017} that arithmetic encoders can trivially be transformed into a recurrent filter, where the output of the modified arithmetic encoder fulfills the \ac{ESP}, i.e., in this case the arithmetic encoder can be interpreted as a reservoir of exactly one unit in the sense of an~\ac{ESN}.

\begin{figure*}[!t]
     \centering
     \begin{subfigure}[t]{0.46\linewidth}
        \centering
        \includegraphics[trim=0 0 0 2mm,clip=true,width=\linewidth]{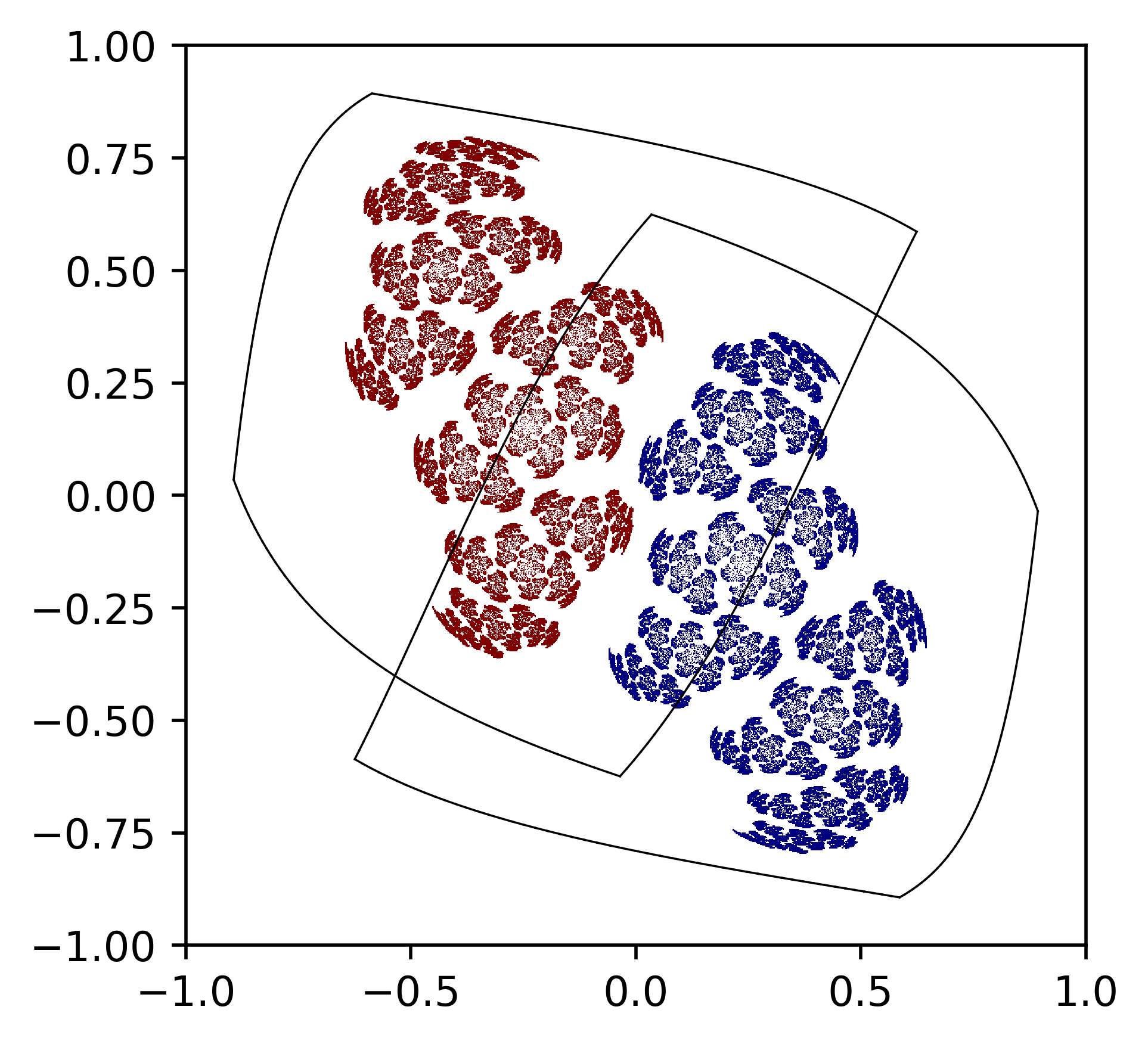}
        \caption{After 1 iteration}
        \label{fig:2d neuron}
     \end{subfigure}
     \hfill
     \begin{subfigure}[t]{0.46\linewidth}
        \centering
        \includegraphics[trim=0 0 0 2mm,clip=true,width=\linewidth]{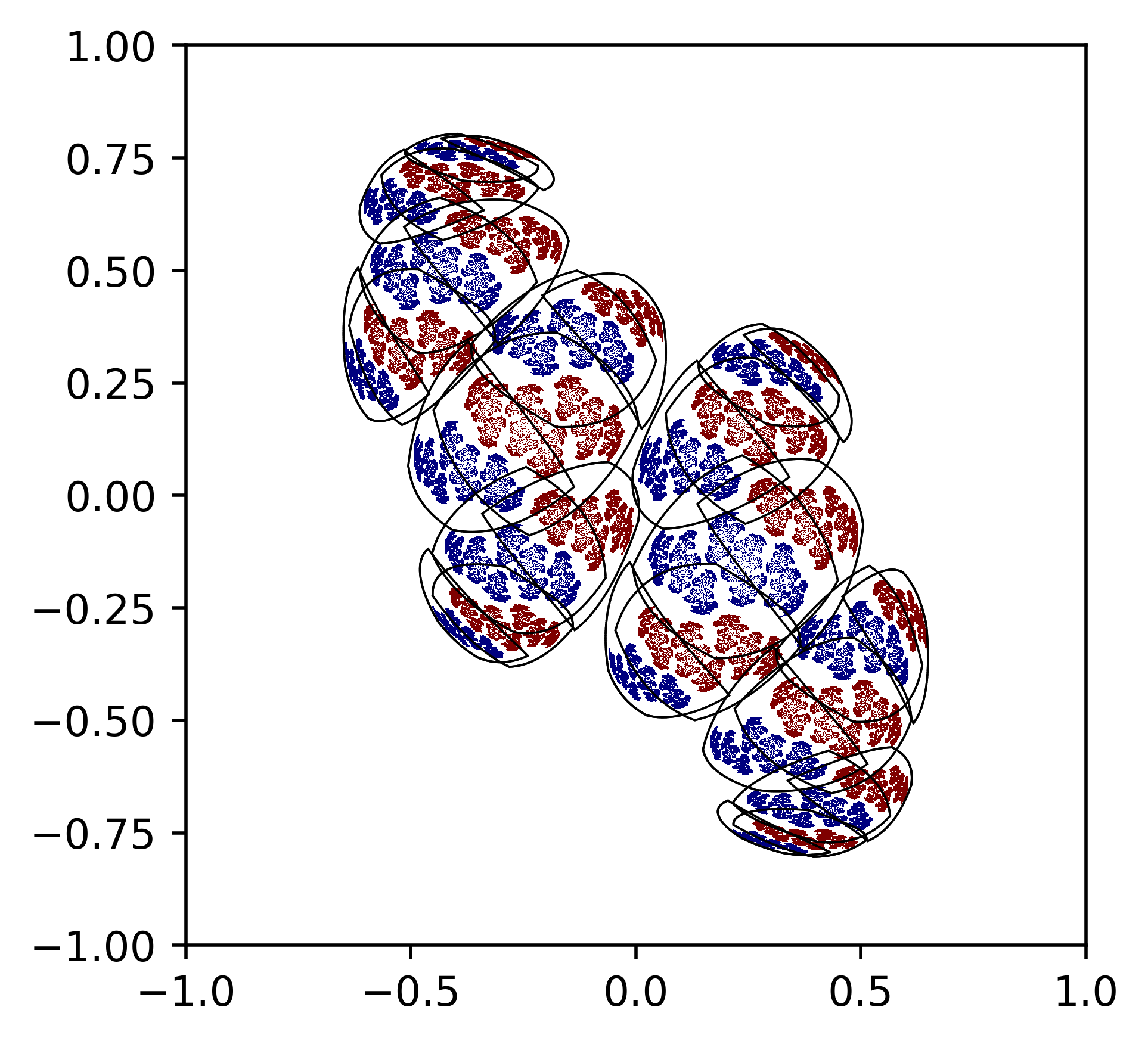}
        \caption{After 5 iterations}
        \label{fig:point cloud}
     \end{subfigure}
     \caption{State space representation of an \ac{ESN}'s recurrent layer with 2 neurons. The closed black curves depict the transformed state space limits.}
 \label{fig:1}
 \end{figure*}

The fractal dimension of a network, as calculated by, e.g., box-counting, is smaller or equal to the number of neurons in the recurrent layer. Before we illustrate the effect of measuring \ac{FD}, we demonstrate the reasons why the \ac{FD} can be a relevant measure for \acp{ESN}:

The graphs in Figs.~\ref{fig:0} and \ref{fig:1} show random state space representations of a recurrent layer with two neurons. The $\tanh$ transfer function restricts the possible values of states to the interval between $-1$ and $1$. For two neurons this defines a square within which all permissible states are located. One can analyze how this square of permissible values is transferred into and transformed for the next iteration. We get two different transformed curves, for the two possible inputs of $1$ and $-1$, at each iteration, cf.~Fig.~\ref{fig:2d neuron}. We also depict representations of different samples of $\mathbb{U}$, where the color of the points depends on whether the last value that the respective time series has received, was, $1$ or $-1$. Fig.~\ref{fig:point cloud} shows the same results of the same network after five iterations.

Although the transformed input spaces overlap in the depicted example 
we do not see areas where the point clouds of different colors inundate each other.

The fractal patterns of Fig.~\ref{fig:0}  are a direct result of the recursive definition of \acp{RNN}. Other known examples of recipes for fractal sets, for example the well known Barnsley fern \cite{barnsley2014fractals}, which is constructed by a recursive rule. In (deep) feedforward networks, FractalNet~\cite{larsson2017fractalnet} creates the architecture of the network based on ideas of self-similarity, but with a different motivation of influencing path lengths for subsequent training, without further investigation of the resulting coding.

Returning to \acp{RNN} and reservoirs, it appears obvious that with a broader coverage of $\mathbb{U}$ within a recurrent layer, information processing capabilities will improve. It might be also worthwhile to consider the maximal information content that is available within the network:
\begin{align*}
   I_{\max} & = m \cdot m_b,
\end{align*}

where $m$ is the number of neurons and $m_b$ is the logarithm of the accuracy where each of the neurons are represented. 
In other words, $I_{\max}$ is the total number of bits available within the reservoir layer, or in $\mathbb{X}$, equivalently.

For example, in the case when values are represented by one double-precision IEEE 754 standard floating-point number, i.e., 
the 64 bits reserved for each number are distributed in the following manner: 1 bit is for the sign, 11 bits for the exponent, and 52 bits for the significant bits (mantissa).
In this example, $m_b = 53$: the number of bits of the mantissa plus the sign bit. For sake of simplicity, we assume here and for the following considerations that the smallest difference (``accuracy'') with which each state can be represented is $2^{-m_b}$. Thus, we neglect the fact when using floating point representations a number that is very near to zero is significantly more accurately represented than numbers with a larger absolute values %
\footnote{%
The concept may be extended to other forms of reservoirs in the sense of RC, in physical systems one may consider the accuracy of reliable readout down to thermal fluctuations, in case of super-cooling systems down to quantum effects, etc.}.

With this as a starting point, we can now imagine the reservoir as a $m$-dimensional hypercube where each side is again subdivided in $2^{m_b}$ segments, forming a total $2^{I_{\max}}$ sub-hyper-cubes. The ideal usage of the reservoir is achieved if all of those sub-hyper-cubes are occupied with a non-empty subset of $\mathbb{U}$ and 
%if those subsets are non-overlapping for two different  hyper-cubes 
are not mapped to the same hypercube for as long as possible lasting input histories 
and the probability for the occurrence of each of the subsets is equal for all hyper-cubes.

Usually, and different from the ideal case, less than the maximal number of those hyper-sub-cubes are actually occupied. With an estimation of \ac{FD}, the number of occupied boxes is 
\[
2^{d_f \cdot m_b}.
\]
A plausible estimate of the mutual information between the input histories and the internal states thus is 
\begin{align*}
I(\mathbb{U};\mathbb{X}) \leq d_f \times m_b,
\end{align*}
 where $d_f$ is the fractal dimension.
 
This suggests we can improve the performance of \acp{ESN} by maximizing the \ac{FD}, by varying the parameters {$\alpha$} and {$\beta$} of the recurrent connectivity. Practically, we need to lift the fractal dimension of the data representation close to the number of neurons such that possible values of the
%%% this sentence needs work, I don't know for sure what it means (OO)
recurrent layer in the \ac{ESN} represent at least one input history. This achieves that the data of the recurrent layer represents a compressed image of an input sequence, for as much as it is possible for the history of $u_t$. Consequently, the representation of the recurrent layer would become something like a Huffman coded representation of the input history, as much as it is possible according to the entropy resulting from the statistics of $\mathbb{U}$.

The basic idea of box counting methods to estimate \ac{FD} of an object is to count the number of boxes $N(\epsilon_i)$ occupied by that object, with boxes at scales $\epsilon_i$. Using a number of different scales $\epsilon_i$, and plotted on a log-log scale, the slope of that resulting line yields the \ac{FD} (``Richardson effect''). 

\begin{align*}
\log N(\epsilon_i) & = d_f \cdot \log\epsilon+b.
\end{align*}

For perfect fractals, this slope is estimated in the limit, but for practical estimation, objects are assumed to be fractal when the slope remains approximately constant over a range of scales. We use a standard box-counting method~\cite{peitgen2006chaos} directly for points which we have obtained from the hidden state output.

\subsubsection{Arithmetic Encoding}
\label{AE:sec}

Arithmetic encoding describes the idea to convert a sequence of symbols into a real-valued number in an iterative procedure. For the encoding, each symbol is assigned an interval with a size proportional to the probability of the occurrence of that particular symbol. In this way, recursively, a sequence of symbols can be encoded until the encoder hits the limit of accuracy of the underlying numerical representation. Here, arithmetic encoding is to map a finite sequence of a discrete set of symbols into a real value $x_t$ in the range$[0,1]$. This real value that represents the sequence can then be transmitted to a receiver. The number of digits transmitted has to be large enough to make it possible to identify the initial sequence. 

Originally, arithmetic encoders have been developed at IBM~\cite{rissanen1979arithmetic} for data compression,  competing with Huffman coding. Interesting  applications of arithmetic encoders can be found in, e.g., \cite{brady1997context,brady2000shape,li2018efficient,li2018enlarging}.  Recent investigations have revealed an analogy between arithmetic encoders and optimal \acp{ESN} \cite{boedecker2012information,boedecker2009studies,mayer2004echo,
mayer2017echo,mayer2017orthogonal,weber2008reservoir}. Making use of this idea, the connectivity in an \ac{RNN} would be arranged such that for the given input statistics in $\mathbb{U}$, the network performs a data compression so that as many as possible past input values are represented in the reservoir.

We rephrase the definition of arithmetic encoders as a recursive filter, and as a result its conformity with the \ac{ESP} becomes obvious\footnote{For this purpose the impact of the time axis has to be inverted from the original approach \cite{rissanen1979arithmetic}.}.
As a pre-requisite, it is necessary to estimate or quantify the probability $p(\kappa_i)$ of a discrete, finite set of symbols $\kappa_i$ with $i \leq I$. 

Now we can define a function $0 \leq g(i) < 1$:
\[g(\kappa_i) = \sum_{j<i}  p(\kappa_j).\]
$g(i)$ is a step-wise monotonously increasing function which can be reverted to 
\[\kappa_i = \tilde{g}(x).\]

To encode a finite sequence $u_t$ from $0 \leq t \leq T$ of those symbols, we choose an initial value $x_0 = a$, where $a$ can be an arbitrary real number in the range $[0,1]$.

We then can calculate recursively for all $0<i \leq T$
\begin{align*}
x_{i+1} = p(u_i)\cdot x_{i} + g(u_i).
\end{align*}

$x_{T+1}$ can then be transmitted. However, this value can only be transmitted using a finite accuracy.
The required accuracy can be derived from considering the initializing setting of $x_0=a$. One can then define
\begin{align*} 
x^{-}      &= x_{T+1} (u_T,\dots,u_0,x_0=0)  \\
x^{+}      &= x_{T+1} (u_T,\dots,u_0,x_0=1)
\end{align*}

The necessary number of digits that have to be transmitted to receive the message unambiguously has to be chosen such that the receiver can know
$ x \in [x^-,x^+] $ but also $x \notin [0, x^-]$ and $x \notin [x^+,1]$. The necessary digits then form an optimally short code, equivalent to a Huffman code.
 
At the same time it is possible to estimate an upper limit to 
\[ x^+-x^- \geq \left(\max_i p(\kappa_i) \right)^{T}. \]
If we assume our time series to have non-zero entropy, we know $\max_i p(\kappa_i)<1$.
Thus, in the limit for $T \to \infty$, this value is zero, i.e., the arithmetic encoder is uniformly state contracting and, as a result, conform with the \ac{ESP}. Arithmetic encoding is a one unit reservoir \ac{ESN}.

Using the arithmetic encoder, with correctly chosen distribution, and if $p(\kappa_i)$ is i.i.d.\ with regard to the position of the symbol within the sequence, the representation of possible symbol sequences is dense on $x_T$, for sufficient large $T$. Thus, the fractal dimension of the mapping of $\mathbb{U}$ on $\mathbb{X}$ is $1.0$. 

The fractal dimension will be lower than $1.0$ if we were using imperfect estimates for $p(\kappa_i)$. As an example, if we were to assume a code statistics of three identical and independent distributed symbols A, B, C with equal probability $\frac{1}{3}$ but in reality the sequences contained only the symbols A and C with equal probability $\frac{1}{2}$, one can organize the structure of the resulting gaps in the corresponding representation as a Cantor set. In this case, we achieve a fractal dimension of $\frac{\log 2}{\log 3} \approx 0.63$.

\section{Results}

In this section we show results of determining the \ac{FD} and subsequent classification using a \ac{SVM}. 
The \ac{SVM} is used to show whether the input sequence mapping to the hidden states of the network is separable. 
 
\begin{figure*}
     \centering
     \begin{subfigure}[t]{0.48\linewidth}
        \centering
        \includegraphics[width=\linewidth]{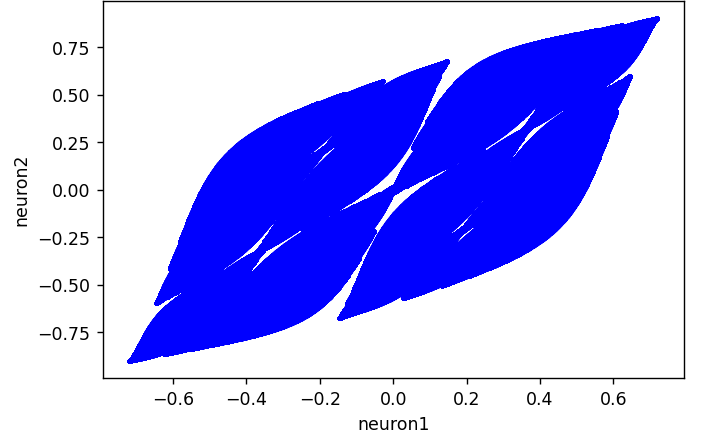}
        \caption{The state space generated by the network for {$\beta=0.45$}. \\
        Plot for $10^8$ sample points on 2 neurons with $\alpha=1$ and $\beta=0.45$.}
        \label{fig:0.45 state space}
     \end{subfigure}
     \hfill
     \begin{subfigure}[t]{0.48\linewidth}
        \centering
        \includegraphics[width=\linewidth]{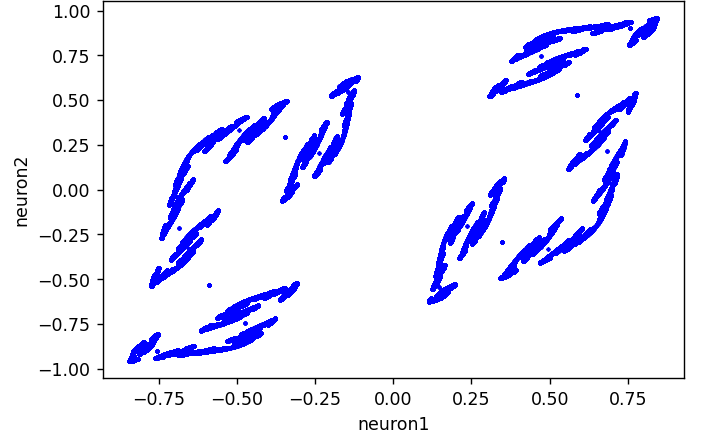}
        \caption{The state space generated by the network for {$\beta=0.8$}. \\
        Plot for $10^8$ sample points on 2 neurons with $\alpha=1$ and $\beta=0.8$.}
        \label{fig:0.8 state space}
     \end{subfigure}
     \begin{subfigure}[t]{0.48\linewidth}
        \centering
        \includegraphics[width=\linewidth]{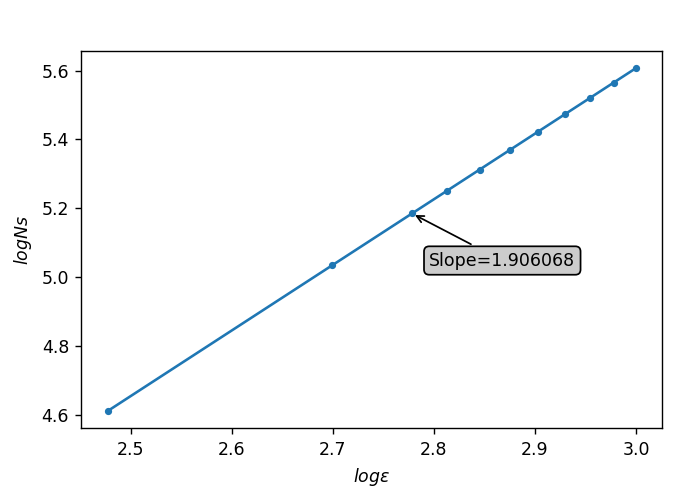}
        \caption{Fractal dimension obtained by Richardson effect for {$\beta=0.45$}.} 
        \label{fig:0.45 box-count}
     \end{subfigure}
     \hfill
     \begin{subfigure}[t]{0.48\linewidth}
        \centering
        \includegraphics[width=\linewidth]{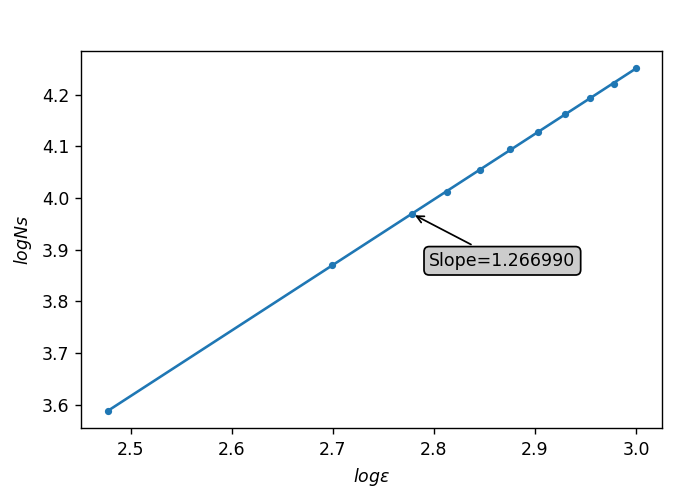}
        \caption{Fractal dimension obtained by Richardson effect for {$\beta=0.8$}.}
        \label{fig:0.8 box-count}
     \end{subfigure}
     \caption{Box-counting results for 2 different beta values.}
 \label{fig:2}
 \end{figure*}

The graphs in Fig.~\ref{fig:2} are result of using the \ac{FD} obtained by applying the box-counting method, where $x$-axis and $y$-axis represents the first neuron and second neuron, respectively. Fig.~\ref{fig:0.45 state space} is the state space visualization generated for {$\alpha=1$} and {$\beta=0.45$}, a promising result. Fig.~\ref{fig:0.8 state space} is an example with poor results for {$\alpha=1$} and {$\beta=0.8$}. For low $\beta$ values, there will not be any gaps between the points, and the goal is to avoid an overlap between points. In this example, $\beta<0.45$ results in overlapping and $\beta>0.45$ results in large gaps between the points as shown in Fig.~\ref{fig:0.8 state space}. Using Richardson effect's box-counting supports this result, see Fig.~\ref{fig:0.45 box-count}. For $\beta=0.45$, we achieve an almost optimal dimensionality of close to 2, equal to the number of neurons. Fig.~\ref{fig:0.8 box-count} shows the unsatisfactory result for $\beta=0.8$. with a dimensionality of 1.2. In Figs.~\ref{fig:0.45 box-count} and \ref{fig:0.8 box-count}, the $x$- and $y$-axes represent the $\log\epsilon$ and $\log N_s$ values,  respectively.

\begin{figure*}
\centering
\begin{minipage}{0.31\linewidth}
\centering
\includegraphics[width=0.98\linewidth]{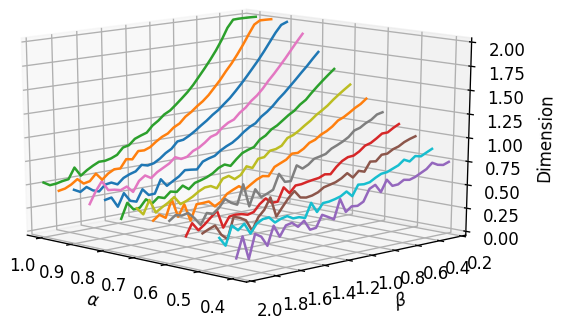}
\caption{Fractal dimension on multiple scales of $\alpha$ and $\beta$.}
\label{fig:3}
\end{minipage}%
\hfill%
\begin{minipage}{0.3\linewidth}
\includegraphics[width=0.95\linewidth]{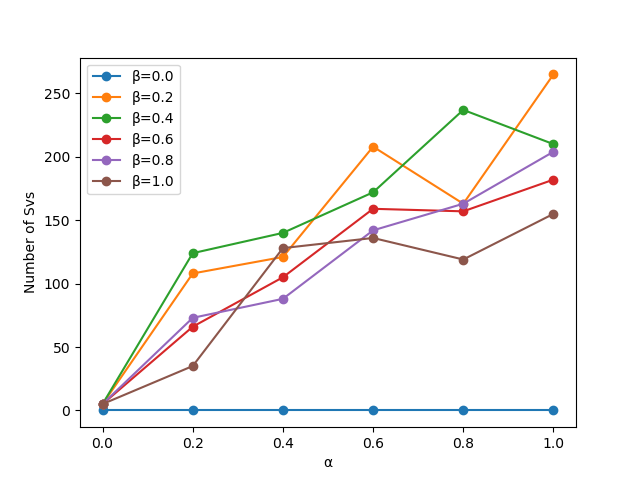}
\caption{Number of support vectors for different $\beta$ and fixed $\alpha$.} % Number of samples: 3000.}
\label{fig:4}
\end{minipage}%
\hfill%
\begin{minipage}{0.3\linewidth}
\includegraphics[width=0.95\linewidth]{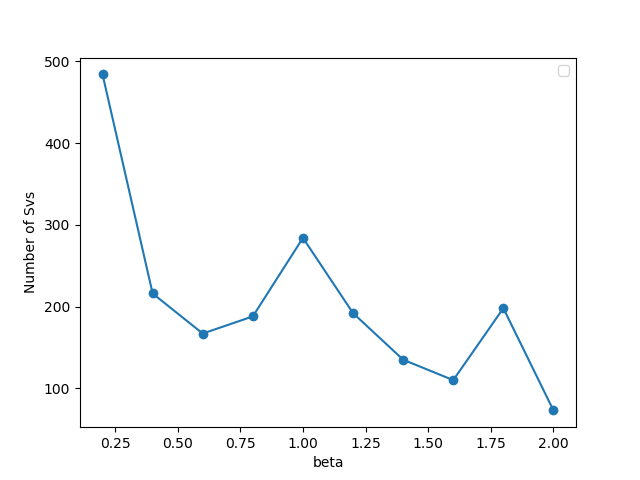}
\caption{Number of support vectors for different $\beta$ and $\alpha=1$.} % Number of samples: 3000.}
\label{fig:5}
\end{minipage}%
\end{figure*}

Fig.~\ref{fig:3} shows a 3d-visualization of \acp{FD} on multiple scales of $\alpha$ and $\beta$. The decrease in $\beta$ is clearly visible as the dimension value increases (the values range from 2.0 to 0.2 from left to right). As $\alpha$ increases, the fractal dimension increases (or vice-versa). That is to say, the measured dimension strongly depends on tuning the $\alpha$ and $\beta$ values.
   
We also observe a trend in the number of support vectors while alpha and beta changes: 
%% I have no idea what this below means (OO)
% also the grammatical input data are added.
The complexity of the hyperplane as generated by the \ac{SVM} model is approximately proportional to the number of support vectors, because each support vector will define one or more twists or turns.

Figure~\ref {fig:4} shows results of a classification with a \ac{SVM}. We plot the number of support vectors for different $\alpha$-values ($x$-axis), with different beta values (indicated by color). When $\alpha$ increases, the number of support vectors has a tendency to increase.

Along with the result of Fig.~\ref{fig:3}, we aim to use the relationship between box-counting and the SVM to analyze \ac{ESN}. When $\alpha=1$ and $\beta=0.2$ or $\beta=0.4$, we achieve to lift the \ac{FD} close to the number of neurons. The same parameters have been used for experiments in both figures.

\subsection{Simulation details}
\label{simulation_details}

For simulations in Fig.~\ref{fig:1} we used $W={\mathcal O}(2.0)$, where $\mathcal{O}$ is a $2\times 2$ rotation matrix around the angle 0.5 rad; $w_\text{in}=\left( \sqrt{0.5}, -\sqrt{0.5} \right)$, $\alpha=0.8$ and $\beta=0.5$. The network has been stimulated by all possible combinations of $1$ and $-1$ up to a limit length emulating iid. random sequences.

For Fig.~\ref{fig:2} and \ref{fig:3}, we used 100 input sequences of 1 and -1, each sequence with about 1 million temporal steps. Hence, for each pair of $\alpha$ and $\beta$, 100 million points were generated in total, using the network with weight matrix
\begin{align*}
W^\text{in}& = \left[
  	\begin{array}{cc}
  	0.8436&       0.7381    \\      
  	\end{array}
  	\right],\, W=
  	\left[
  	\begin{array}{cc}
  	   0.0169&     0.5711    \\\nonumber
  	   1.2895&     0.2509      
  	\end{array}
  	\right],
\end{align*}

Moreover, the interval of $\alpha$ and $\beta$ were set to 
\begin{align*}
\begin{split}
\boldsymbol\alpha:[0.4,0.45,0.5\cdots,1]  &= \{X\in \mathbb{R}|x_n-x_{n-1}=0.5\};\\
\boldsymbol\beta:[0.15,0.2,0.25\cdots,2]  & = \{X\in \mathbb{R}|x_n-x_{n-1}=0.5\}
\end{split}
\end{align*}

The slope and the 3D illustration in Fig.~\ref{fig:2} and Fig.~\ref{fig:3}, respectively, were produced by the box-counting method with
\begin{equation*}
\epsilon=\{300, 500, 600, 650, 700, 750, 800, 850, 900, 950 ,1000\}\nonumber
\end{equation*}
The simulation of Fig. \ref{fig:4} consists of 3000 training data generated from network states in SVM model. The weight matrix used for Fig.~\ref{fig:4} was same as the one used for Fig.~\ref{fig:2}. The values $\alpha$ and $\beta$ are tuned between 0 and 1.
The classifier to explore separation is a radial basis function kernel SVM, with the Gaussian kernel:
\begin{eqnarray}
K(\mathbf{x_a}, \mathbf{x_b}) = exp\Bigg( - \frac{ \parallel \mathbf{x_a} - \mathbf{x_b} \parallel ^2 }{2 \sigma^2} \Bigg)\nonumber
\end{eqnarray}
where $\sigma$ is a tunable parameter.  This parameter influences the ``reach'' of each single example: when $\sigma$  is small, the SVM model can capture more complex data. We chose $\sigma = 0.05$,  to have a good ability of classifying the overlapping samples, which means the shape of decision boundary may be more complicated. An additional hyper-parameter $C$ trades off correctness and complexity of the model (essentially a regularization parameter). We set $C= 2.0$.

\section{Discussion}

One one hand, the arithmetic encoding approach has a strong relationship to \acp{ESN} with one neuron, and complies with the \ac{ESP}. One the other hand, it is well established that arithmetic encoders are optimal encoders of an (input) sequence. In the ideal case, for correctly applied arithmetic encoders and appropriate input sequences, this will lead to a compact, dense state representation of input sequences, resulting in a \ac{FD} of 1 (per neuron). % (?)
% Is 1.0 correct? Or is it 1 for each neuron. Should it be the number of neurons rather than 1?
At the same time overlapping representations have to be avoided. Heuristics hint towards the notion that both conditions are fulfilled exactly at the limit point where the \ac{FD} reaches the reservoir size.

For input sequences composed from discrete sets, we see arithmetic encoders as a very valuable guideline to understand optimal representations in reservoirs, leading to new criteria for reservoir and \ac{RNN} initialization. 
We suggest the following new concepts with regard to optimizing reservoirs:
\begin{itemize}
\item The \ac{FD} of the mapping $\mathbb{U} \rightarrow \mathbb{X}$ shall be near the number of neurons. This an important requirement, since the requirement of the complete coverage of the layer of input neurons appears to be difficult in the case of the \ac{ESN} formulation.
\item The modified \ac{ESN} shall work like a non-linear vector arithmetic encoder.
\item Reservoirs with overlapping representations have indistinguishable states for readouts by means of regression or classification. We have to avoid those overlapping representations of different input histories. 
\end{itemize}

This result has implications for a previously formalized idea of power law forgetting~\cite{neco2015} and the edge of chaos heuristics: for input sequences with non-zero entropy, this idea would lead to overlapping representations. Overlapping representations of several recently variant input time series have clearly an impact on the performance of all ESNs including those with hundreds of recurrent neurons. Since this phenomenon occurs at a lower bound than the heuristically found edge of chaos it seems plausible that future discussions might lead to a new bound for the ESP, that is the ESP of non-overlapping representations with regard to a given input statistics.

\section{Conclusion}
The fractal structure of representations in \acp{RNN} is to our best knowledge an unnoticed aspect with implications for reservoir computing. It opens a new field of research which can bring together aspects of fractal analysis, information theory, and representation learning. Our results have been combined with the insight that overlapping representations of variant input time series better shall be avoided. This leads us to still preliminary proposed ESP II, for which a less heuristic and better analytically founded formulation is still ongoing work in progress.

\section{Acknowledgment}
We thank Bo Ruei Jiang, Gorri Anil Kumar, and Ming Jie Li for their help. We thank Advanced Institute of Manufacturing with High-tech Innovations (AIM-HI) for their financial support.

\bibliography{references}

\begin{thebibliography}{10}
\providecommand{\url}[1]{#1}
\csname url@samestyle\endcsname
\providecommand{\newblock}{\relax}
\providecommand{\bibinfo}[2]{#2}
\providecommand{\BIBentrySTDinterwordspacing}{\spaceskip=0pt\relax}
\providecommand{\BIBentryALTinterwordstretchfactor}{4}
\providecommand{\BIBentryALTinterwordspacing}{\spaceskip=\fontdimen2\font plus
\BIBentryALTinterwordstretchfactor\fontdimen3\font minus
  \fontdimen4\font\relax}
\providecommand{\BIBforeignlanguage}[2]{{%
\expandafter\ifx\csname l@#1\endcsname\relax
\typeout{** WARNING: IEEEtran.bst: No hyphenation pattern has been}%
\typeout{** loaded for the language `#1'. Using the pattern for}%
\typeout{** the default language instead.}%
\else
\language=\csname l@#1\endcsname
\fi
#2}}
\providecommand{\BIBdecl}{\relax}
\BIBdecl

\bibitem{jaeger2001echo}
H.~Jaeger, ``The ``echo state'' approach to analysing and training recurrent
  neural networks -- with an erratum note,'' German National Research Centre
  for Information Technology, GMD report 148, 2001.

\bibitem{jaeger2002short}
H.~Jaeger, ``Short term memory in echo state networks,'' German National
  Research Centre for Information Technology, GMD report 152, 2002.

\bibitem{jaeger2002tutorial}
H.~Jaeger, ``Tutorial on training recurrent neural networks, covering {BPPT},
  {RTRL}, {EKF} and the ``echo state network'' approach,'' German National
  Research Centre for Information Technology, GMD report 159, 2002.

\bibitem{WLS04}
O.~L. White, D.~D. Lee, and H.~Sompolinsky, ``Short-term memory in orthogonal
  neural networks,'' \emph{Phys. Rev. Lett.}, vol.~92, p. 148102, Apr 2004.

\bibitem{mayer2017orthogonal}
N.~M. Mayer and Y.-H. Yu, ``Orthogonal echo state networks and stochastic
  evaluations of likelihoods,'' \emph{Cognitive Computation}, vol.~9, no.~3,
  pp. 379--390, 2017.

\bibitem{boedecker2012information}
J.~Boedecker, O.~Obst, J.~T. Lizier, N.~M. Mayer, and M.~Asada, ``Information
  processing in echo state networks at the edge of chaos,'' \emph{Theory in
  Biosciences}, vol. 131, no.~3, pp. 205--213, 2012.

\bibitem{yildiz2012re}
I.~B. Yildiz, H.~Jaeger, and S.~J. Kiebel, ``Re-visiting the echo state
  property,'' \emph{Neural networks}, vol.~35, pp. 1--9, 2012.

\bibitem{buehner2006tighter}
M.~Buehner and P.~Young, ``A tighter bound for the echo state property,''
  \emph{IEEE Transactions on Neural Networks}, vol.~17, no.~3, pp. 820--824,
  2006.

\bibitem{mayer2004echo}
N.~M. Mayer and M.~Browne, ``Echo state networks and self-prediction,'' in
  \emph{International workshop on biologically inspired approaches to advanced
  information technology}.\hskip 1em plus 0.5em minus 0.4em\relax Springer,
  2004, pp. 40--48.

\bibitem{hajnal2006critical}
M.~A. Hajnal and A.~L{\H{o}}rincz, ``Critical echo state networks,'' in
  \emph{International Conference on Artificial Neural Networks}.\hskip 1em plus
  0.5em minus 0.4em\relax Springer, 2006, pp. 658--667.

\bibitem{pentland1984fractal}
A.~P. Pentland, ``Fractal-based description of natural scenes,'' \emph{IEEE
  transactions on pattern analysis and machine intelligence}, no.~6, pp.
  661--674, 1984.

\bibitem{peitgen2006chaos}
H.-O. Peitgen, H.~J{\"u}rgens, and D.~Saupe, \emph{Chaos and fractals: new
  frontiers of science}.\hskip 1em plus 0.5em minus 0.4em\relax Springer
  Science \& Business Media, 2006.

\bibitem{balghonaim1998maximum}
A.~S. Balghonaim and J.~M. Keller, ``A maximum likelihood estimate for
  two-variable fractal surface,'' \emph{IEEE Transactions on Image Processing},
  vol.~7, no.~12, pp. 1746--1753, 1998.

\bibitem{gagnepain1986fractal}
J.~Gagnepain and C.~Roques-Carmes, ``Fractal approach to two-dimensional and
  three-dimensional surface roughness,'' \emph{wear}, vol. 109, no. 1-4, pp.
  119--126, 1986.

\bibitem{sarkar1994efficient}
N.~Sarkar and B.~B. Chaudhuri, ``An efficient differential box-counting
  approach to compute fractal dimension of image,'' \emph{IEEE Transactions on
  systems, man, and cybernetics}, vol.~24, no.~1, pp. 115--120, 1994.

\bibitem{chen2003two}
W.-S. Chen, S.-Y. Yuan, and C.-M. Hsieh, ``Two algorithms to estimate fractal
  dimension of gray-level images,'' \emph{Optical Engineering}, vol.~42, no.~8,
  pp. 2452--2465, 2003.

\bibitem{novianto2003near}
S.~Novianto, Y.~Suzuki, and J.~Maeda, ``Near optimum estimation of local
  fractal dimension for image segmentation,'' \emph{Pattern Recognition
  Letters}, vol.~24, no. 1-3, pp. 365--374, 2003.

\bibitem{jaeger2007echo}
H.~Jaeger, ``Echo state network,'' \emph{scholarpedia}, vol.~2, no.~9, p. 2330,
  2007.

\bibitem{jaeger2005reservoir}
H.~Jaeger, ``Reservoir riddles: Suggestions for echo state network research,''
  in \emph{Proceedings. 2005 IEEE International Joint Conference on Neural
  Networks, 2005.}, vol.~3.\hskip 1em plus 0.5em minus 0.4em\relax IEEE, 2005,
  pp. 1460--1462.

\bibitem{xue2007decoupled}
Y.~Xue, L.~Yang, and S.~Haykin, ``Decoupled echo state networks with lateral
  inhibition,'' \emph{Neural Networks}, vol.~20, no.~3, pp. 365--376, 2007.

\bibitem{steil2007online}
J.~J. Steil, ``Online reservoir adaptation by intrinsic plasticity for
  backpropagation--decorrelation and echo state learning,'' \emph{Neural
  Networks}, vol.~20, no.~3, pp. 353--364, 2007.

\bibitem{song2010stable}
Q.~Song, Z.~Feng, and M.~Lei, ``Stable training method for echo state networks
  with output feedbacks,'' in \emph{2010 International Conference on
  Networking, Sensing and Control (ICNSC)}.\hskip 1em plus 0.5em minus
  0.4em\relax IEEE, 2010, pp. 159--164.

\bibitem{liu2009improved}
Y.~Liu, J.~Zhao, W.~Wang, Y.~Wu, and W.~Chen, ``Improved echo state network
  based on data-driven and its application to prediction of blast furnace gas
  output,'' \emph{Acta Automatica Sinica}, vol.~35, no.~6, pp. 731--738, 2009.

\bibitem{obst2010iconip}
O.~Obst, J.~Boedecker, and M.~Asada, ``Improving recurrent neural network
  performance using transfer entropy.'' in \emph{ICONIP 2010: Neural
  Information Processing, Models and Applications}, 2010, pp. 193--200.

\bibitem{sheng2012prediction}
C.~Sheng, J.~Zhao, Y.~Liu, and W.~Wang, ``Prediction for noisy nonlinear time
  series by echo state network based on dual estimation,''
  \emph{Neurocomputing}, vol.~82, pp. 186--195, 2012.

\bibitem{rissanen1979arithmetic}
J.~Rissanen and G.~G. Langdon, ``Arithmetic coding,'' \emph{IBM Journal of
  research and development}, vol.~23, no.~2, pp. 149--162, 1979.

\bibitem{BC2017}
\BIBentryALTinterwordspacing
N.~M. Mayer, ``Arithmetic encoding, reservoir computing, criticality and
  biological implications,'' in \emph{Bernstein Conference, G\"ottingen}, 2017,
  (Abstract). [Online]. Available:
  \url{https://abstracts.g-node.org/abstracts/bfcd6994-61b6-4b25-83bb-e3ce10217745}
\BIBentrySTDinterwordspacing

\bibitem{barnsley2014fractals}
M.~F. Barnsley, \emph{Fractals everywhere}.\hskip 1em plus 0.5em minus
  0.4em\relax Academic press, 2014.

\bibitem{larsson2017fractalnet}
G.~Larsson, M.~Maire, and G.~Shakhnarovich, ``{FractalNet}: Ultra-deep neural
  networks without residuals,'' in \emph{ICLR}, 2017.

\bibitem{brady1997context}
N.~Brady, F.~Bossen, and N.~Murphy, ``Context-based arithmetic encoding of 2d
  shape sequences,'' in \emph{Proceedings of International Conference on Image
  Processing}, vol.~1.\hskip 1em plus 0.5em minus 0.4em\relax IEEE, 1997, pp.
  29--32.

\bibitem{brady2000shape}
N.~Brady and F.~Bossen, ``Shape compression of moving objects using
  context-based arithmetic encoding,'' \emph{Signal Processing: Image
  Communication}, vol.~15, no. 7-8, pp. 601--617, 2000.

\bibitem{li2018efficient}
M.~Li, S.~Gu, D.~Zhang, and W.~Zuo, ``Efficient trimmed convolutional
  arithmetic encoding for lossless image compression,'' \emph{arXiv preprint
  arXiv:1801.04662}, pp. 107--120, 2018.

\bibitem{li2018enlarging}
M.~Li, S.~Gu, D.~Zhang, and W.~Zuo, ``Enlarging context with low cost:
  Efficient arithmetic coding with trimmed convolution,'' \emph{arXiv preprint
  arXiv:1801.04662}, 2018.

\bibitem{boedecker2009studies}
J.~Boedecker, O.~Obst, N.~M. Mayer, and M.~Asada, ``Studies on reservoir
  initialization and dynamics shaping in echo state networks.'' in
  \emph{ESANN}, 2009.

\bibitem{mayer2017echo}
N.~M. Mayer, ``Echo state condition at the critical point,'' \emph{Entropy},
  vol.~19, no.~1, p.~3, 2017.

\bibitem{weber2008reservoir}
C.~Weber, K.~Masui, N.~M. Mayer, J.~Triesch, and M.~Asada, ``Reservoir
  computing for sensory prediction and classification in adaptive agents,''
  \emph{Machine Learning Research Progress: Nova publishers: Hauppauge, NY,
  USA}, 2008.

\bibitem{neco2015}
N.~M. Mayer, ``Input-anticipating critical reservoirs show power law forgetting
  of unexpected input events,'' \emph{Neural Computation, MIT Press}, vol.~27,
  pp. 1102--1119, May 2015.

\end{thebibliography}
\bibliographystyle{IEEEtran}

\end{document}